\documentclass[sigconf,authorversion,nonacm]{acmart}
\usepackage{amsmath}

\usepackage{amssymb}
\usepackage{graphicx}
\graphicspath{ {.} }

\begin{document}

%
\title{A framework for massive scale personalized promotion}

%

\author{Yitao Shen}
\authornote{Both authors contributed equally to this research.}
\email{yitao.syt@antgroup.com}
\affiliation{%
 \institution{Ant Financial Services Group}
 \city{Hangzhou}
 \state{Zhejiang}
 \country{China}
}

\author{Yue Wang}
\authornotemark[1]
\email{ben.wy@antgroup.com}
\affiliation{
 \institution{Ant Financial Services Group}
 \city{Hangzhou}
 \state{Zhejiang}
 \country{China}
}

\author{Xingyu Lu}
\email{sing.lxy@antgroup.com}
\affiliation{
 \institution{Ant Financial Services Group}
 \city{Hangzhou}
 \state{Zhejiang}
 \country{China}
}

\author{Feng Qi}
\email{feng.qi@antgroup.com}
\affiliation{%
 \institution{Ant Financial Services Group}
 \city{San Mateo}
 \state{California}
 \country{United States}
}

\author{Jia Yan}
\email{yanjia.jy@antgroup.com}
\affiliation{%
 \institution{Ant Financial Services Group}
 \city{Shanghai}
 \country{China}
}

\author{Yixiang Mu}
\email{yixiang.myx@antgroup.com}
\affiliation{
 \institution{Ant Financial Services Group}
 \city{Hangzhou}
 \state{Zhejiang}
 \country{China}
}

\author{Yao Yang}
\email{yangyao.yy@antgroup.com}
\affiliation{
 \institution{Ant Financial Services Group}
 \city{Hangzhou}
 \state{Zhejiang}
 \country{China}
}

\author{Yifan Peng}
\email{ifan@std.uestc.edu.cn}
\affiliation{
 \department{School of Computer Science and  engineering}
 \institution{University of Electronic Science and Technology of China}
 \city{ChengDu}
 \state{SiChuan}
 \country{China}
}

\author{Jinjie Gu}
\authornote{Corresponding author}
\email{jinjie.gujj@antgroup.com}
\affiliation{%
 \institution{Ant Financial Services Group}
 \city{Hangzhou}
 \state{Zhejiang}
 \country{China}
}

\begin{abstract}
Technology companies building consumer-facing platforms may have access to massive-scale user population. In recent years, promotion with quantifiable incentive has become a popular approach for increasing active users on such platforms. On one hand, increased user activities can introduce network effect, bring in advertisement audience, and produce other benefits. On the other hand, massive-scale promotion causes massive cost. Therefore making promotion campaigns efficient in terms of return-on-investment (ROI) is of great interest to many companies.

This paper proposes a practical two-stage framework that can optimize the ROI of various massive-scale promotion campaigns. In the first stage, users' personal promotion-response curves are modeled by machine learning techniques. In the second stage, business objectives and resource constraints are formulated into an optimization problem, the decision variables of which are how much incentive to give to each user. In order to do effective optimization in the second stage, counterfactual prediction and noise-reduction are essential for the first stage. We leverage existing counterfactual prediction techniques to correct treatment bias in data. We also introduce a novel deep neural network (DNN) architecture, the deep-isotonic-promotion-network (DIPN), to reduce noise in the promotion response curves. The DIPN architecture incorporates our prior knowledge of response curve shape, by enforcing isotonicity and smoothness. It out-performed regular DNN and other state-of-the-art shape-constrained models in our experiments.

\end{abstract}

%
%


%

%
\maketitle

\section{Introduction}

\begin{figure}
\includegraphics[width=0.5\textwidth]{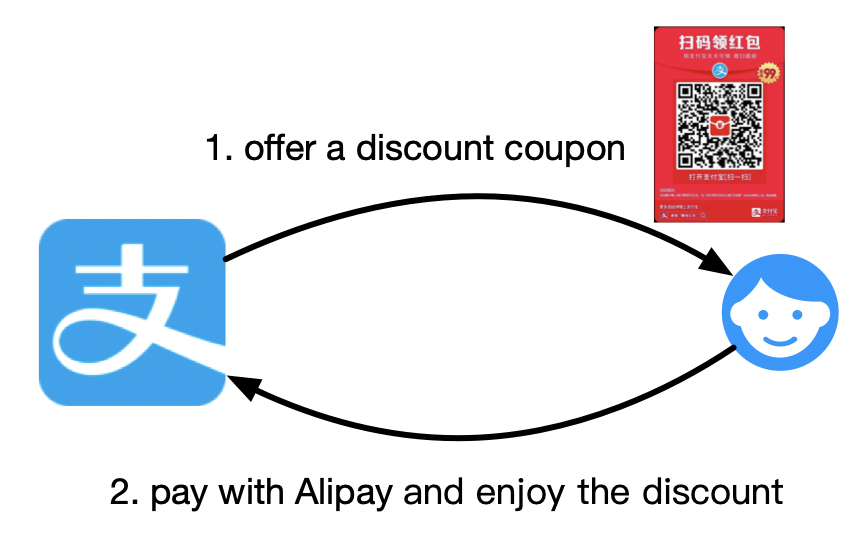}
\caption{illustration of the Alipay marketing campaign.}
\label{biz}
\end{figure}

Digital platforms nowadays serve various demands of societies, for example, e-commerce, ride sharing, and personal finance. For-profit platforms have strong motivation to grow the sizes of their active user bases, because larger active user bases introduce more network effect, more advertisement audience, more cash deposit, etc.

To convert inactive users to active users, one established way is to use personalized recommendation systems \cite{youtubeRecsys}\cite{netflixRecsys}. A platform can infer users' personal interests from profile information and behavior data, and recommend content accordingly. Recommendation systems rely on good product design, "big data", efficient machine learning algorithms\cite{itemCF}, and high-quality engineering systems \cite{amatriain_basilico_2016}.

Recently, the approach of using promotion incentive, such as coupon, to convert users has become popular \cite{CHRISTINO201978}\cite{Li2019LatentDA}. To enjoy the incentive, users are required to finish specified tasks, for example subscribing to a service, purchasing a product, sharing promotional information on a social network, etc. The key decision to make in such promotion campaigns is how much incentive to give to each user.

Our work is in the hypothetical context of Alipay offline payment marketing campaign, though it can be easily generalized to other incentive promotion campaigns. Gaining offline payment market share is a main business objective of Alipay. Ren-chuan-ren-hong-bao (social network based red packet) is the largest marketing campaign hosted by Alipay to achieve this goal. In this campaign, Alipay granted coupons to customers to incentivize them to make mobile payments with the Alipay mobile app. Given its marketing campaign budget, the company needed to determine the value of the coupon given to each user to maximize overall user adoption. We illustrate the marketing campaign in \autoref{biz}.

We propose a two-stage framework for solving the personalized incentive decision problem. In the first stage, we model users' personal promotion-response curves with machine learning algorithms.
In the second stage, we formulate the problem as a linear programming (LP) problem and solve it by established LP algorithms.

\begin{figure}
\includegraphics[width=0.5\textwidth]{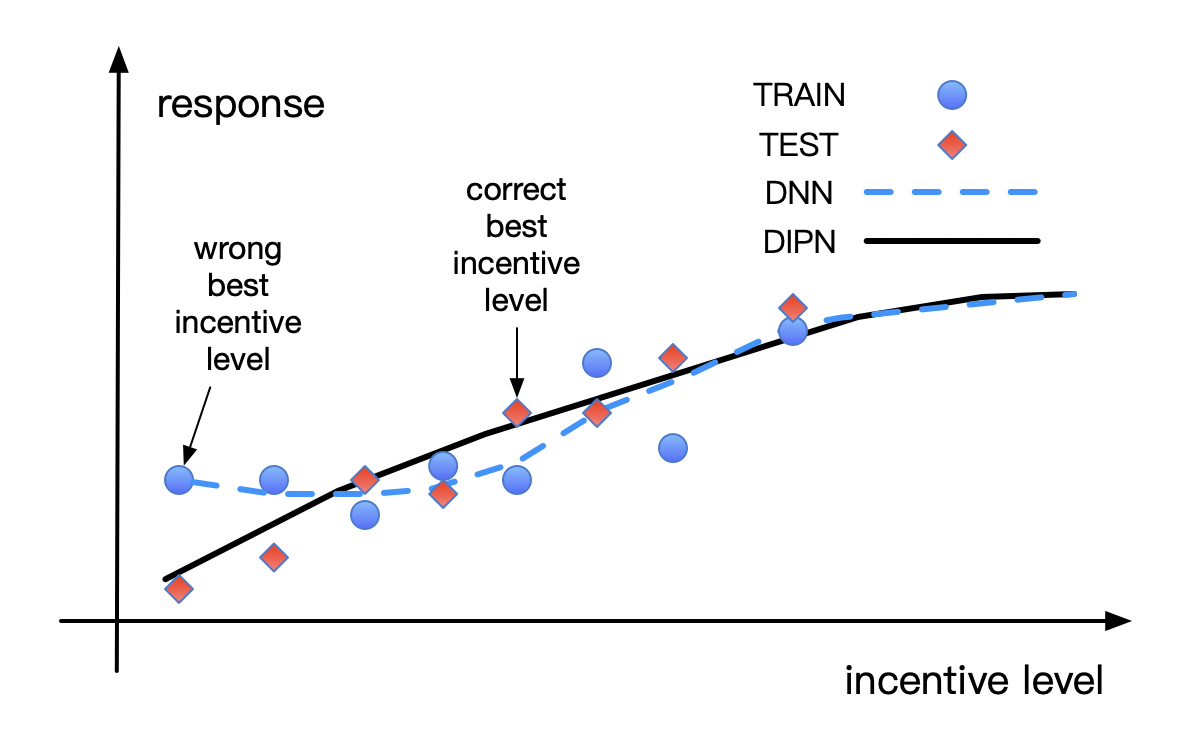}
\caption{An illustration of how noise in data causes a sub-optimal incentive decision, and DIPN is more immune to noise compared to DNN. Rounds and diamonds represent average response rates at different promotion incentive (cost) levels in training and testing data, respectively. The amount of training data at low cost levels is small and hence the average response rate is noisy. The estimation by DNN based on the training data indicates that a low incentive should be applied (marked as "wrong best cost"). When this decision is made, a large amount of users receive incentive at low levels, and a large amount of "testing data" become available, reducing the noise and revealing the suboptimality of the original decision. DIPN is shape-constrained and more immune to the low training data quality. Based on DIPN, correct decision can still be made.}
\label{demo}
\end{figure}

In practice, modeling promotion-response curves is challenging due to data sparsity and noise. Real-world promotion-response datasets usually lack samples for certain incentive values, even though the total amount of samples is large. Such sparsity combined with noise causes inaccuracy in response curve modeling and sub-optimal decision in incentives. We introduce a novel isotonic neural network architecture, the deep-isotonic-promotion-network (DIPN), to alleviate this problem. DIPN incorporates our prior knowledge of the response curve shape by enforcing isotonicity and regularizing for smoothness. It out-performed regular DNN and other state-of-the-art shape-constrained models in our experiments. \autoref{demo} illustrates such an example. 

\begin{figure}
\includegraphics[width=0.5\textwidth]{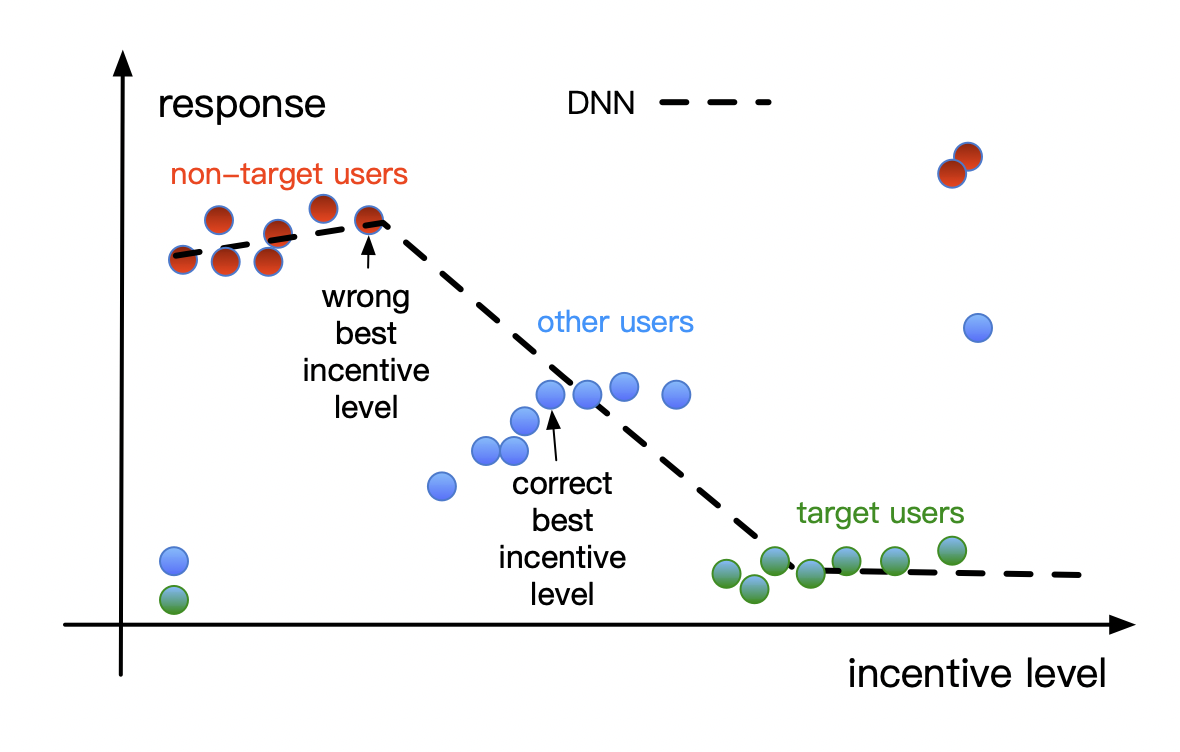}
\caption{Illustration of how treatment bias in training dataset leads to wrong incentive decisions. There are three types of users: active (red), ordinary (blue), and inactive (green). Each type has higher response rate to higher incentive (cost). However, in collected data, user activity is negatively correlated with incentive. A DNN model without knowledge of this bias will follow the dotted line, which does not reflect the causal effect of incentive. Based on this DNN model, a decision engine will never use any incentive larger than the "wrong best cost".}
\label{demo2}
\end{figure}

Another well known challenge for response curve modeling is treatment bias in historical data. If data samples are not collected through randomized trials, naively fitting  the relationship between incentive and response cannot capture the true causal effect of incentive.  \autoref{demo2} illustrates an example on how a biased  dataset causes sub-optimal incentive decision. In real-world marketing campaigns, collecting fully randomized incentive samples is cost-ineffective, because it means randomly giving a large amount of users random amount of incentives. We use the inverse propensity score (IPS) \cite{austin2011introduction} technique to correct treatment bias.

\section{Related work}

\subsection{Large-scale or personalized optimization problems based on prediction}
\label{sec:combining-prediction-and-optimization}
Many companies, e.g. Linkedin, Pinterest, Yahoo!, have developed solutions for large-scale optimization based on predictive models.  \cite{Xu:2015:SPE:2783258.2788615} focused on online advertisement pacing. The authors developed user response prediction models and optimized for performance goals under budget constraints. 
\cite{agarwal2012personalized} focused on online content recommendation. The authors developed personalized click shaping plans by solving a LP problem in its dual space. \cite{Gupta:2016:MCI:2946645.3007062}\cite{Gupta:2017:OEV:3132847.3132849}\cite{Zhao:2018:NVC:3219819.3219906} study the problem of email volume control. The predicted responses at different levels of email volume served as coefficients of engagement maximization problems. To our knowledge, there are few published studies on large-scale personalized incentive optimization.

\subsection{Causal inference and counterfactual prediction}
 Incentive response modeling should establish causal relationship between incentive and user response. The model is used in a counterfactual way to answer the question "what if users are given incentive levels different from that is observed". There is a large volume of literature on  counterfactual modeling, e.g. \cite{bonner2018causal}\cite{hartford2016counterfactual}\cite{swaminathan2015counterfactual}. We use inverse propensity score (IPS) \cite{austin2011introduction}\cite{rosenbaum1983central} to weight sample data when the data collection process is not randomized. This is assumed in our  framework unless otherwise stated.

\subsection{Shape-constrained models}
Shape constraints include constraints on monotonicity and convexity (concavity). They are a form of regularization, introducing bias and prior knowledge. Shape constraints are helpful in below scenarios.
\begin{itemize}
\item Monotonicity or convexity (concavity) is desired for interpretability. For example, in economy theory, marginal gain on increasing promotion incentive should be positive but diminishing \cite{kahneman2013prospect}. In our experience, promotion response is non-decreasing, but the marginal gain does not necessarily decrease. We thus propose to apply just monotonicity constraint to promotion response models.
\item Prior knowledge on function shapes exists, but training data is too sparse to guarantee such shapes without regularization. In our experience, this is usually true for promotion response modeling, because a reasonable model is required as early as possible after a promotion campaign kicks off, allowing very limited time for training data collection. At the same time, we do know that response rate should monotonically increase with incentive.
\end{itemize}

The related works section of \cite{gupta2018diminishing} summarized four categories of shape-constrained models:

\begin{itemize}
\item General additive models (GAMs). A GAM is a summation of multiple one dimensional functions. Each of the 1-d function takes one input feature, and is responsible for enforcing the desired shape for that input.
\item Max-affine functions, which express piece-wise linear convex functions as the max of a set of affine functions. If the derivative with respect to an input is restricted to be positive/negative, monotonicity can also be enforced.
\item Monotonic neural networks. Neural networks can be viewed as recursive functions, with the output of one layer serving as the next layer's input. For a recursive function to be convex increasing, it is sufficient if its input function and the function itself are convex increasing. For a recursive function to be convex, it is sufficient if its input function is convex and the recursion is convex increasing.
\item Lattice networks \cite{You:2017:DLN:3294996.3295058}. The simplest form of lattice network is linear interpolation built on a grid of input space. Monotonicity and convexity are enforced as linear constraints on first and second order derivatives when solving for the model. The grid dimension grows exponentially with the input dimension, so the authors ensembled multiple low-dimension lattice networks built on different subsets of inputs to handle high dimensional data.
\end{itemize}

\cite{gupta2018diminishing} showed that lattice network has more expressive power than monotonic neural network since the former allows convex and concave inputs to coexist, and that lattice network is at least as good as monotonic neural network in accuracy.

Our work proposes the DIPN architecture (discussed in \autoref{isotonic}) that constrains NN to be monotonic for one input i.e. the incentive level. It does not aim to compete with state-of-the-art shape constrained models in terms of expressiveness, but instead aim to provide high accuracy and interpretability for promotion response modeling.

\section{The personalized promotion framework}
\label{sec_pnp}
The two steps of our framework for making personalized incentive decisions are (1) incentive-response modeling and (2) user response maximization under incentive budget. As the second step is the decision making step, and the first step prepares coefficients for it, we start \autoref{sec_pnp} by describing the optimization problem in the second step assuming a incentive-response model is already available, and dive deep into our approach for the incentive-response model in \autoref{isotonic}.

\autoref{tab:math_symbol_meaning} summarizes mathematical notations necessary for this section.

\begin{table}[h!] 
  \caption{Notations}
  \label{tab:math_symbol_meaning}
  \begin{center}
    \begin{tabular}{c|p{0.8\columnwidth}} 
      symbol&meaning\\
      \hline
      $x_i$& feature vector of user i\\
      $c_i$& incentive for user i \\     
      $y_i$& response label for user i \\     
      $f(x, c)$& user response prediction function with two inputs: x is the user feature vector, and c is the incentive \\ 
      $g_k(x, c)$& user cost prediction function for the k-th resource \\
      $d_j$& the j-th incentive level bin after discretizing the incentive space,$d_j < d_{j+1}, \forall j$ \\
      $D$& total number of incentive level bins after discretizing the incentive space\\
       $z_{ij}$ & decision variable representing the probability of choosing $d_j$ for the i-th user \\
       $B$& total campaign budget \\
       $b$& budget per capita\\
    \end{tabular}
  \end{center}
\end{table}

The objective of our formulation is to maximize total future user responses, and the constraint is the limited budget. Here future response can an arbitrarily defined business metric, e.g. click-through rate or long-term user value.

\begin{equation} \label{nonlinear}
\begin{split}
       &\max_{c_i}\sum_{i}f(x_i, c_i) \\
  s.t. &\sum_ic_i\leq B
\end{split}
\end{equation}

\subsection{Solving the optimization problem}
Because the number of users is large, and $f(x_i, c_i)$ is usually nonlinear and non-convex, \autoref{nonlinear} can be difficult to solve. We can restrict promotion incentive $c_i$ to a fixed number of $D$ levels: $\{{d_j}| j=0,1, ... D-1\}$, and use assignment probability variables $z_{ij} \in [0,1]$ to turn \autoref{nonlinear} into an LP.

\begin{equation} \label{mip_lp}
\begin{split}
       &\max_{z_{ij}}\sum_i\sum_jf(x_i, d_j)z_{ij} \\
  s.t. &\sum_{i}\sum_{j}d_jz_{ij} \leq B \\
       &\sum_{j}z_{ij}=1, \forall i \\
       & z_{ij} \in [0, 1], \forall i, j
\end{split}
\end{equation}

Solution $z_{ij}$ of \autoref{mip_lp} should be on one of the vertices of its feasible polytope, hence most of the values of $z_{ij}$ should be $0$ or $1$. For fractional $z_{ij}$ in solutions, we treat $z_{ij}, \forall i$ as a probability simplex and sample from it. This is usually good enough for production usage. 

\autoref{mip_lp} can be solved by commercial solvers implementing off-the-shelf algorithms such as primal-dual or simplex. If the number of users is  too large, \autoref{mip_lp} can be solved by dual ascent \cite{boyd2011distributed}. Note the dual problem of the LP can be decomposed into many user-level optimization problems and solved in parallel. A few specialized large-scale algorithms can also be applied to problems with the structure of \autoref{mip_lp}, e.g. \cite{zhong2015stock}\cite{zhang2020solving}.

One advantage of \autoref{mip_lp} is that $f(x_i, d_j)$ is computed before solving the optimization problem. Hence the specific choice of the functional form of $f(x_i, d_j)$ does not affect the difficulty of the optimization problem. Whether $f(x_i, d_j)$ is a logistic regression model or a DNN model is transparent to \autoref{mip_lp}.

Sometimes promotion campaigns have more than one constraints. A commonly seen example is that overlapped subgroups of users are subject to separate budget constraints. In general, we can model any constraints by predictive modeling, and then formulate a multiple-constraint problem. Suppose for each user $i$ there are $K$ kinds of costs modeled by:

\begin{equation} \label{g_k}
\begin{split}
  g_k(x_i, c_i), k=0,1,.. K-1
\end{split}
\end{equation}

The multiple-constraint optimization problem is:

\begin{equation} \label{multi-lp}
\begin{split}
 &\max_{z_{ij}}\sum_{i}\sum_{j}f(x_i, d_j)z_{ij} \\
  s.t. &\sum_{i}\sum_{j}g_k(x_i, d_j)z_{ij} \leq B_k, k=0,1,... K - 1 \\
       &\sum_{j}z_{ij}=1, \forall i \\
       & z_{ij} \in [0, 1], \forall i, j
\end{split}
\end{equation}

A frequently seen alternative to constrain the total budget $B$ is to constrain budget per capita $b$. The corresponding formulation is as below.
\begin{equation} \label{avg_lp}
\begin{split}
 &\max_{z_{ij}}\sum_{i}\sum_{j}f(x_i, d_j)z_{ij} \\
s.t. &\sum_{i}\sum_{j}(g_k(x_i, d_j)-b_k)z_{ij}\leq 0, \forall k \\
       &\sum_{j}z_{ij}=1, \forall i \\
       & z_{ij} \in \{0, 1\}, \forall i, j
\end{split}
\end{equation}

\autoref{multi-lp} and \autoref{avg_lp} are both LPs since $f(x_i, d_j)$ and $g_k(x_i, d_j)$ are pre-computed coefficients. They can be solved by the same solvers used for \autoref{mip_lp}.

\subsection{Online decision making for new users}
Sometimes it is desired to be able to make incentive decisions for a stream of incoming users. It is not possible to put such users together and solve one optimization problem beforehand. However, if we can assume the dual variables is stable over a short period of time, such as one hour, we can solve the optimization problem with users in the previous hour, and reuse the optimal dual variables in the next hour. This assumption is usually not applicable to general optimization problems, but when the amount of users is large and user population is stable, it holds. This approach can also be viewed from the perspective of shadow price \cite{boyd2004convex}. The Lagrangian multiplier can be interpreted as the marginal objective gain when one more unit of budget is available. This marginal gain should not change rapidly for a stable user population.

We thus break down the optimization step into a dual variable solving step and a decision making step. The decision making step can make incentive decision for a single user. Consider the dual formulation of \autoref{mip_lp}, and let $\lambda$ be the Lagrangian multiplier for the budget constraint:
\begin{equation} \label{dual}
\begin{split}
       &\min_{\lambda}\max_{z_{ij}}\sum_i\sum_jf(x_i, d_j)z_{ij} - \lambda(\sum_{i}\sum_{j}d_jz_{ij} - B) \\
       s.t.
       &\sum_{j}z_{ij}=1, \forall i \\ 
       & z_{ij} \in [0, 1], \forall i, j \\
       & \lambda > 0
\end{split}
\end{equation}

If $\lambda$ is given, \autoref{dual} can be decomposed into a per user optimization policy:

\begin{equation} \label{individual}
\begin{split}
&\max_{z_j}(f(x_i, d_j) - \lambda d_j)z_{ij}\\
       s.t.
       &\sum_{j}z_{ij}=1, \forall i \\ 
       & z_{ij} \in [0, 1], \forall i, j \\
\end{split}
\end{equation}

\autoref{individual} is applicable to unseen users as long as $x_i$ is known.

\subsection{Challenges for two-stage framework}
"User's optimal incentive level" is the lowest level for a certain user to be converted. If we offer less incentive, we will lose a user conversion(situation A). If we offer more incentive, we waste a certain amount of marketing funds.

\begin{figure*}

\begin{minipage}{0.49\textwidth}
\includegraphics[width=0.9\textwidth]{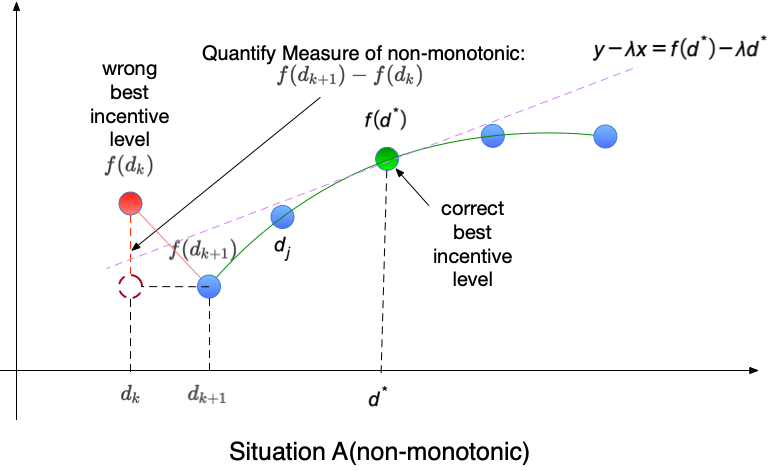}
\label{situationA}
\end{minipage}
\hfill
\begin{minipage}{0.49\textwidth}
\includegraphics[width=0.9\textwidth]{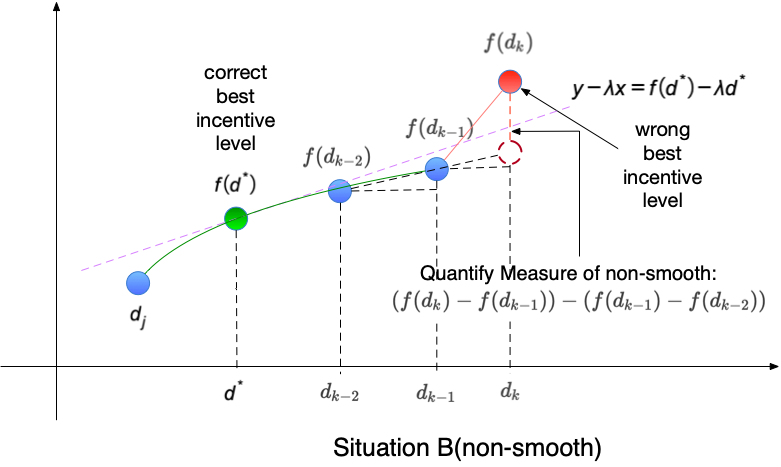}
\label{situationB}
\end{minipage}
\caption{An illustration of two common bad cases in real world datasets, $d^*$ is the expected optimal incentive level, if $f(d_j)$ satisfy $f(d_j)-\lambda d_j < f(d^*)-\lambda d^*, \forall d_j \neq d^*$, our framework outputs the right answer $d^*$. But we often found some predicted score like $f(d_k), f(d_k) - \lambda d_k > f(d^*)-\lambda d^*$, become a wrong best incentive level.
}
\label{situation}
\end{figure*}

For illustration, we could relax constraint by supposing that optimal $z$ is on its $[0, 1]$ boundary, omit the notation of index $i$ and rewrite formula \ref{individual} as
\begin{equation} \label{individual_rewrite}
\begin{split}
&\max_{z_j}r(d_j, y_j)\\
       s.t.
       &y_j=f(x, d_j) \\
       &\sum_{j}z_{j}=1  \\ 
       & z_{j} \in [0, 1], \forall j \\
\end{split}
\end{equation}
where $r(d, y)=y - \lambda d$.

\autoref{demo} illustrates wrong prediction leads to wrong best incentive level. In this section, \autoref{situation} illustrates two situations of them in more details. Considering $d^*$ is the optimal incentive level for user $i$, according to \autoref{individual}, we hope our prediction function satisfy $f(d_j)-\lambda d_j < f(d^*)-\lambda d^*, \forall d_j \neq d^*$. If due to lack of data, $f(d_k)-\lambda d_k > f(d^*)-\lambda d^*$, incentive level $d_k \neq d^*$ will be chosen, and a wrong decision will be made.

\vspace{6pt}
\noindent\textbf{Situation A: non-monotonic}

the user response prediction function frequently gives a high prediction $f(d_k)$ at a lower incentive level $d_k$. And because $d_k < d^*$, user $i$ didn't get a satisfactory incentive, we will lose a customer.

In this situation, we introduce prior knowledge to constrain the response curve's shape: users' expected response monotonically increases with the promotion incentive. Therefore, $f(d_{k+1}) - f(d_k)$ must greater than zero.
\vspace{6pt}

\noindent\textbf{Situation B: non-smooth}

In other situations, a very high prediction $f(d_k)$ is given at a higher incentive level $d_k$.  And because $d_k > d^*$, $d_k - d^*$ marketing funds is wasted.

In situation B, we introduce another prior knowledge to constrain the response curve's shape: the incentive-response curves are smooth. Therefore, $(f(d_{k}) - f(d_{k-1})) - (f(d_{k-1}) - f(d_{k-2}))$ should not be too large.

\vspace{10pt}
Based on the discussion of these two situations above, we introduce a novel deep-learning structure DIPN to avoid both situations as far as possible.

\section{Deep isotonic promotion network (DIPN)} \label{isotonic}
\begin{figure}
\centering
\includegraphics[width=0.5\textwidth]{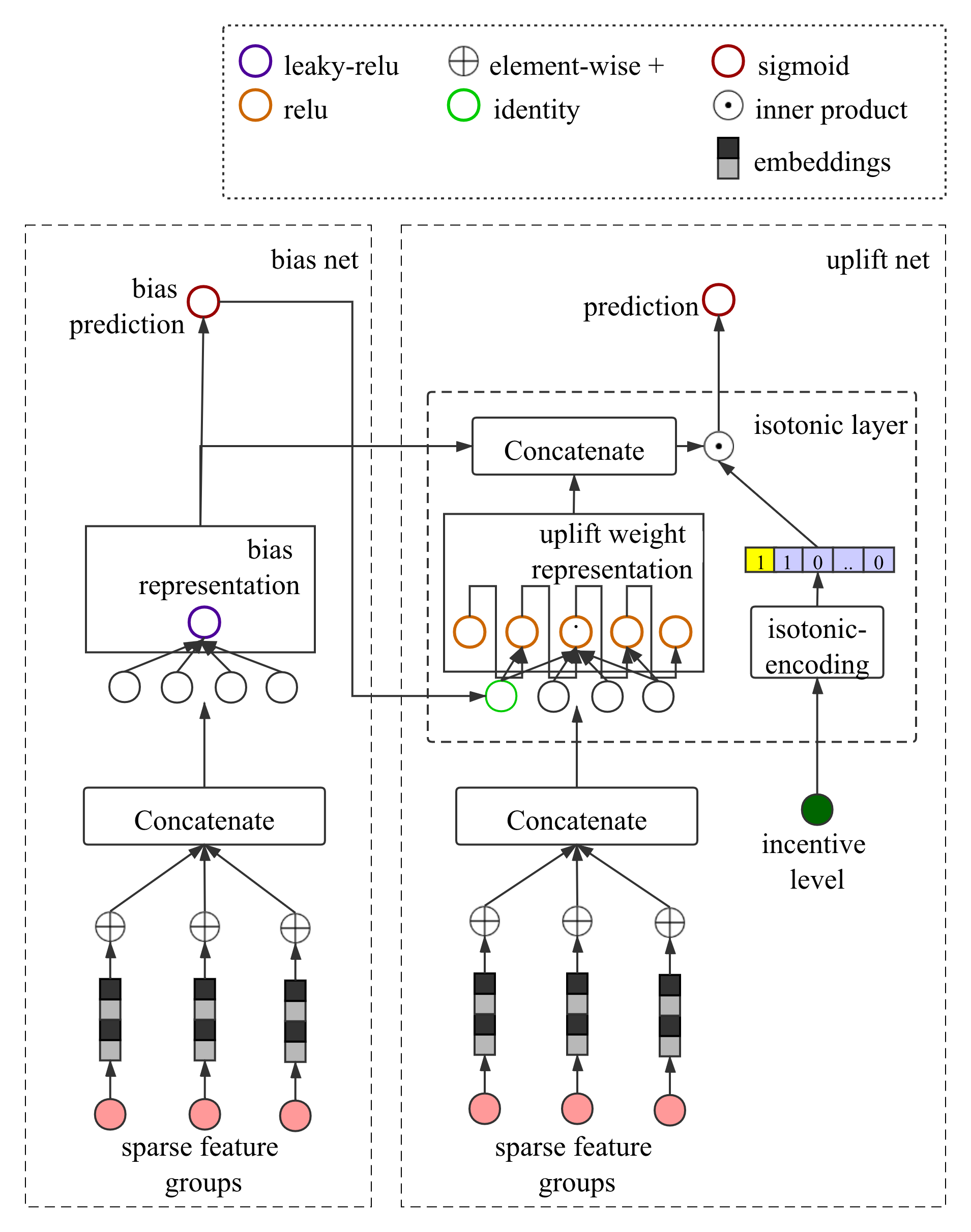}
\caption{The architecture of DIPN. DIPN composes of bias net and uplift net as shown on the left and right sides
. In both nets, sparse user features are one-hot encoded and mapped to their embedding, which are aggregated by summation. Then in bias net, concatenated sparse features are feed to one fully connected layer whose one dimension output activated by leaky ReLU is treated as logit of bias prediction. The bias prediction is inputted to uplift net. In uplift net, another fully connected layer takes concatenated sparse features as input and outputs  $D$ ReLU activated positive values as uplift weights. The previous uplift weight will be inputted to the subsequent node for generating next uplift weight. The isotonic layer outputs the uplift weight representation $w(x)$.}
\label{dipnfig}
\end{figure}
In practice, promotion events often do not last for long, so there is business interest to serve promotion-response models as soon as possible after a campaign starts. Given limited time accumulating training data, incorporating prior knowledge to facilitate modeling is desirable. We choose to enforce the promotion response to be monotonically increasing with incentive level.

DIPN is a DNN model designed for learning user promotion-response curve. DIPN predicts the response value for a given discretized incentive level and a user's feature vector. We can get a user's response curve by enumerating all incentive levels. The response curve learned by DIPN satisfies both monotonicity and smoothness. DIPN achieves this using its isotonic layer (discussed later). Incentive level, which is a one-dimensional discrete scalar, and user features are all inputs to the isotonic layer. While incentive level is inputted to the isotonic layer directly, user features can be transformed by other layers. DIPN consists of bias net and uplift net. The term uplift refers to response increment due to incentive increment. While prediction result of the bias net gives the user's response estimate for minimum incentive, the uplift net learns the uplift response. The DIPN architecture is shown in \autoref{dipnfig}. In the remaining of \autoref{isotonic}, we focus on explaining the isotonic layer and learning process.

\subsection{Isotonic Embedding} \label{Isotonic embedding}
\begin{figure}
\centering
\includegraphics[width=0.48\textwidth]{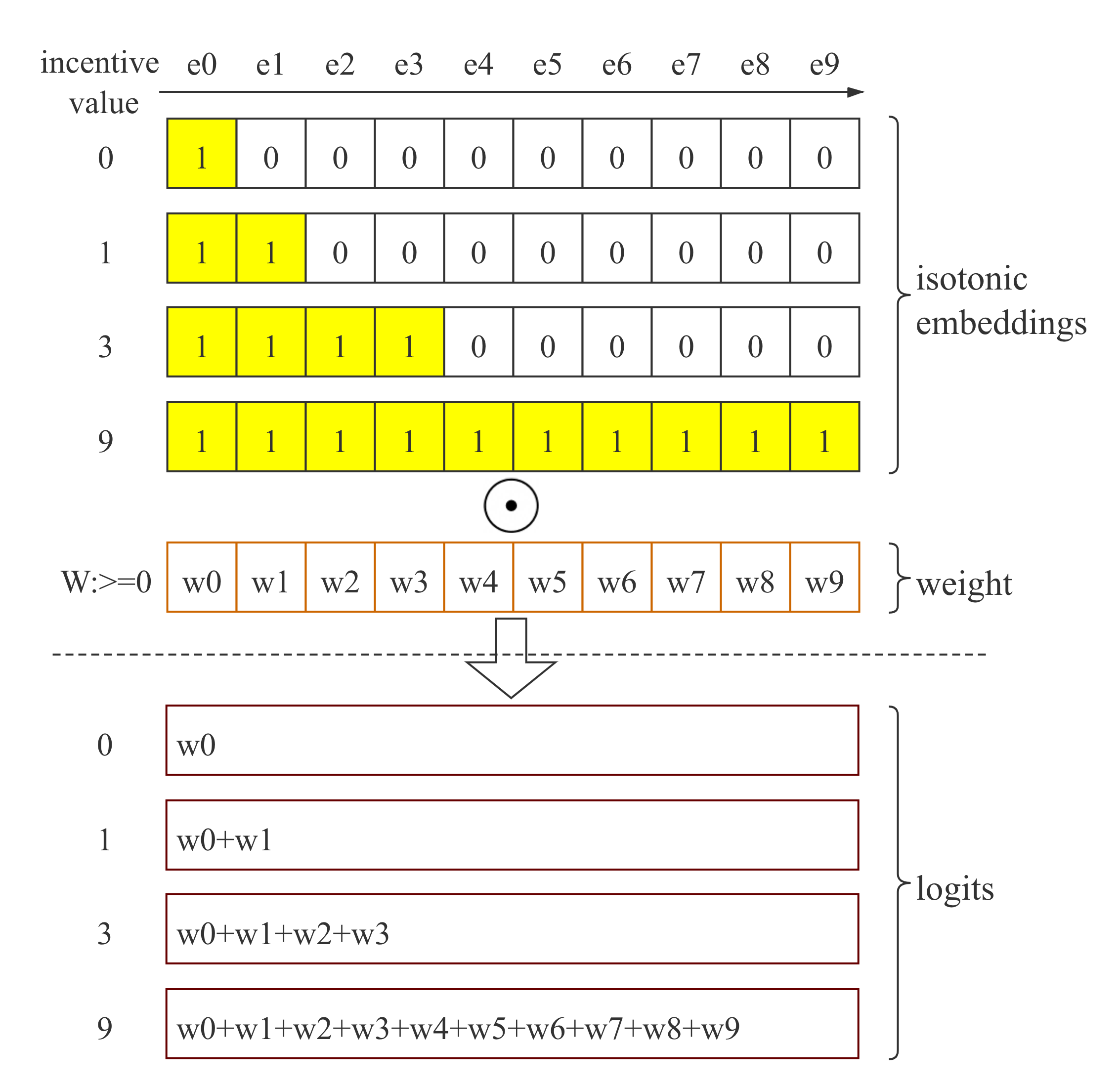}
\caption{Logistic regression with isotonic embedding}
\label{ielrfig}
\end{figure}
Isotonic embedding transforms an incentive level to a vector of binary values. Each digit in the vector represents one level. All digits representing levels lower than or equal to the input level are ones. We use $e(c) \in \{0, 1\}^{D}$ to denote the $D$-digit isotonic embedding of incentive level $c$, and $d_j$ to denote the incentive level of the $j-th$ digit, thus 
\begin{equation} 
\begin{split}
  & e_j(c)=
\begin{cases}
    1,& \text{if } c\geq  d_j\\
    0,& \text{otherwise}
\label{ielr}
\end{cases}
\end{split}
\end{equation}

If we fit a logistic regression response curve with non-negative weights using isotonic embedding as input, the resulting curve will be monotonically increasing with incentive.  Several examples are given in \autoref{ielrfig}. 

\begin{equation} 
\begin{split}
  f(c) &= sigmoid(\sum_j w_je_j(c)+b) \\
       &= {\frac{1}{(1 + exp(- \sum_j w_je_j(c)-b))}} \\
  s.t. &w_j\geq 0, \forall j\\
\label{ielr}
\end{split}
\end{equation}

It is trivial to show the monotonicity of $f(c)$ since $sigmoid$ function is monotonic, and 
\begin{equation} 
\begin{split}
&(\sum_j w_je_j(c + \triangle c) + b) - (\sum_j w_je_j(c)  + b) \\   
&= \sum_{d_j=c}^{d_j=c+\triangle c}{w_j} \geq 0, \forall \triangle c \ge 0
\label{monotone-proving}
\end{split}
\end{equation}

\subsection{Uplift Weight Representation} \label{Uplift weight}
In DIPN, the non-negative weights are output of DNN layers. These weights are thus personalized. We name the personalized non-negative weight the uplift weight, because it is proportion to the uplift value, defined as the incremental response value corresponding to one unit increase in incentive value.

In a binary classification scenario, the prediction function of DIPN is:
\begin{equation}
\begin{split}
& f(x,c)=sigmoid(\sum_j w_j(x)e_j(c)+b)
\end{split}
\label{dipn-prediction-function}
\end{equation}
where $w_j(x)$ is the uplift weight representation,  and $e_j(c)$ is the isotonic embedding. $w_j(x)$ is learned by a neural network using ReLU activation function in last layer so that $w_j(x) $ is non negative . 

 For two consecutive incentive levels $d_{j}$ and $d_{j+1}$, $w_{j+1}$ is a uplift measure for the incremental treatment effect of increasing incentive from $d_{j}$ to $d_{j+1}$. To see this, approximate $sigmoid$ function by its first order expansion around $d_j$:

\begin{equation} 
\begin{split}
f(d_{j+1}) -  f(d_{j})  &= g(z_{j+1}) - g(z_{j}) \\ 
&\simeq g'(z_{j})(z_{j+1} - z_{j}) \\
&=g'(z_{j})w_{j+1} 
\label{firstorder}
\end{split}
\end{equation}
where $g$ is $sigmoid$ function and  $g'$ is its first order derivative, $z_j = \sum_j w_je_j(d_j)+b$. Hence in a small region surrounding $d_{j}$, if $g'(z_{j})$ can be seen as a constant,  $w_{j+1}$ is proportional to the uplift value at $f(d_j)$.

\subsection{Smoothness} \label{Monotonicity and smoothness}
Smoothness means that response value does not change much as the incentive level varies. Users' incentive-response curves are usually smooth when fitted with sufficient amount of unbiased data. We therefore add regularization in the DIPN loss function to enforce smoothness.

\begin{equation}
L=\frac{1}{M}\sum_i{log\_loss(f(x_i, c_i), y_i) +\alpha \cdot smoothness\_loss(w(x_i))}
\label{dipn-loss-function}
\end{equation}
where $M$ is the number of training data points,  $log\_loss$ measures the degree of fitting to training data, $smoothness\_loss$ measures smoothness of predicted response curve, and $\alpha > 0$ balances the two losses. 

The definition of $log\_loss$ and the definition of $smoothness\_loss$ are given in \autoref{log-loss-def} and \autoref{smoothness-loss-def} respectively. User index $i$ in \autoref{log-loss-def} and user feature vector $x_i$ in \autoref{smoothness-loss-def} are omitted for simplicity.
\begin{equation}
log\_loss(f(x, c), y) = -y log(f(x, c)) - (1 - y) log(1 - f(x, c))
\label{log-loss-def}
\end{equation}\begin{equation}
smoothness\_loss(w) = \frac{1}{D}\sum_{j}{(w_{j+1} - w_{j}) ^ 2 / (w_{j+1} w_{j})}
\label{smoothness-loss-def}
\end{equation}

One necessary and sufficient condition of smoothness of the predicted response curve is the uplift values of consecutive incentive levels are as close as enough. As we have proven  in \autoref{Uplift weight} that the uplift value can be approximated by the uplift weight representation, we want the difference of the uplift weights $(w_{j+1} - w_{j})^2$ to be small enough.  $(w_{j+1} w_{j})$ is added to denominator of $smoothness\_loss$ to normalize the  differences over all incentive levels. 

\subsection{Two Phases Learning} \label{Two Phases Learning}
For stability, the training process is split into two phases-bias net learning phase (BLP) and uplift net learning phase (ULP). 

BLP learns the bias net while ULP learns the uplift net. In BLP, only samples with lowest incentive are used for training, and only bias net is activated which is easy to converge. 

In ULP, all samples are used and variables in bias net are fixed. The fixed bias net sets up a robust initial boundary for uplift net learning. Bias net's prediction will be inputted to isotonic layer to enhance uplift net's capability based on the consumption that users with similar bias are more likely to have similar uplift responses. 

Another difference in UWLP is smoothness loss's weight $\alpha$. $\alpha$ will be decaying gradually in UWLP as we observed that larger $\alpha$ helped model converging faster at the beginning of training. The logloss will dominates total loss eventually. $\alpha$'s updating formula is given as follow:

\begin{equation}
\begin{split}
&\alpha=max(\alpha^{l}, \alpha^{u}- \gamma \cdot global\_step)
\end{split}
\label{decayed-smoothness-loss-weight}
\end{equation}
where $\alpha^{u}$ is initial upper bound of $\alpha$, $\alpha^{l}$ is final lower bound of $\alpha$, $\gamma$ controls decaying speed.

\section{Experiment}
In experiments, we focus on comparing DIPN with other response models, solving the same optimization problems \autoref{mip_lp} with the same budget. Specifically, we compare (1) regular DNN, (2) ensemble Lattice network \cite{You:2017:DLN:3294996.3295058}, and DIPN. For each model, we search for a good architecture configuration and hyperparameter setting, and calculate below metrics:
\begin{itemize}
\item Logloss: Measures the likelihood of the fitted model to explain the observations.
\item AUC-ROC: Measures the correctness of the predicted probability order of being positive events.
\item Reverse pair rate (RPR): For each observation of user-incentive-response in the dataset, we make counterfactual predictions on the response of the same user given all possible incentives. For any pair of two incentives, if incentive $a$ is larger than $b$, and the predicted response of $a$ is smaller than that of $b$, we find one reversed pair. For $n$ incentives, there are $\frac{n(n-1)}{2}$ pairs. If there are $rp$ reversed pairs, the RPR is defined as $\frac{2rp}{n(n-1)}$. RPR can be viewed as the degree of violating the response model monotonicity. To obtain RPR for a population, we average all users' RPR values.
\item Equal pair rate (EPR): Similar to RPR, but instead of counting the pairs of reversed pairs, EPR counts the pairs having equal predicted response. We consider low EPR as a good indicator for monotonicity. 
\item Max local slope standard deviation (MLSS): Similar to RPR, for each user, we make counterfactual prediction on response for each incentive. For every two consecutive incentives $c_1$ and $c_2$, assuming their predicted responses are $p_1$ and $p_2$, we can compute the local slope $s_1=\frac{p_1-p_2}{c_1-c_2}$. Consider a range on incentive $[c_i - r, c_i + r]$, we collect all local slopes inside this range, compute their standard deviation. Across all such incentive ranges, we use the maximum local slope standard deviation as MLSS. This metric reflects smoothness of the response curve. Average of all users MLSS is used as a model's MLSS.
\item Future response: Applying the strategy learned from training data to a new user population, following \autoref{individual}, the average response rates of this population. Higher response rate is better. With synthetic datasets, for which we know the ground truth data generation process, we can use the true response expectation on the promotion given by the strategy to evaluate the response outcome. With real-world datasets, we can search in the holdout dataset for the same type of user with the same promotion given by the strategy, and consider all such users will show the observed response on this promotion. This approach can be viewed as importance sampling.
\item Future cost error: Similar to future response, applying the strategy learned from the training data to a holdout population, the user-average cost may exceed the business constraint. We can follow the evaluation method for the future response metric, instead of compute the response rates, compute the response induced cost that exceeds budget.
\end{itemize}

\subsection{Synthetic and Production datasets}
We use synthetic dataset, for which we know the ground truth of data generation process, to evaluate our promotion solution. The feature space consists of three 1-in-n categorical variables, with $n_1$, $n_2$, and $n_3$ categories, respectively. The joint distribution of the three categories consists of $n_1n_2n_3$ different combinations. For each combination, we randomly generate four parameters $a \sim U(0, 1)$, $b = 1 - a$, $\mu \sim U(-50, 150)$, and $\delta \sim U(0, 50)$, where $U(l, u)$ is the uniform distribution bounded by $l$ and $u$. A curve $y=f(x)$ is then generated as follows:
\begin{flalign} 
y = a + b\int_0^x \exp{-\frac{(t-\mu)^2}{2\delta^2}} dt, x\in[0, 100]
\label{gen}
\end{flalign}

The discrete version is:
\begin{flalign} 
&y[i] = a + \frac{b}{100Z}\sum_{h=0}^i \exp{-\frac{(h-\mu)^2}{2\delta^2}}, i=0,1,..100 \\
&Z = \max_h\exp{-\frac{(h-\mu)^2}{2\delta^2}}, h=0,1,..100
\label{gen_discrete}
\end{flalign}

The curves of $y=f(x)$ is used as the ground truth of the expected response $y$ on different incentive $x$, for each joint category of users. Without loss of generality, we constrain the incentive range to be $[0, 100]$. It's easy to see that the curve is monotonically increasing, convex if $\mu < 0$, and concave if $\mu > 100$.

To generate pseudo dataset with noise and and sparsity, we first generate a random integer $z$ between 1 and 1000 with equal probability, for each feature combination. With $z$ being the number of data points for this combination, we randomly choose a promotion integer $p$ between 1 and 100 with equal probability for each data point. With promotion $p$ and expected response $y = f(p)$, a $0/1$ label is generated with probability $y$ of being 1.

On this dataset, we generated 20000 data points, 5000 training samples, 5000 validation samples and 10000 testing samples.

The evaluation metrics for the response models are shown below tables. \autoref{tab:table1} shows the simulation results on synthetic data with $n_1=2, n_2=2, n_3=2$. DIPN showed the highest future response rate and lowest future cost. \autoref{tab:table2} shows the simulation results on synthetic data with $n_3=2, n_2=5, n_3=7$. Again, DIPN showed the highest future response rate and lowest future cost. \autoref{tab:table3} shows the results on a real promotion campaign, DIPN showed the lowest future cost and the same future response rate as that of DNN. Overall DIPN consistently outperformed other models in our test.

\begin{table}[h!]
  \begin{center}
  \caption{Response model evaluation on synthetic data 1($n_1=2, n_2=2, n_3=2$)}
    \label{tab:table1}
    \begin{tabular}{c|c|c|c} 
      &\textbf{DNN} & \textbf{Lattice} & \textbf{DIPN}\\
      \hline
      LogLoss(lower is better) & 0.5713 & 0.5772 & 0.5770 \\
      AUC-ROC & 0.6976 & 0.6931 & 0.6967 \\
      RPR(lower is better) & 0.153 & 0 & 0\\
      EPR(lower is better) & 0 & 0.048 & 0.001\\
      MLSS(lower is better) & 0.003 & 0.007 & 0.006\\
      Future response & 0.743 & 0.710 & 0.759\\
      Future cost($constraint=11.0$) & 11.0 & 11.7$(*)$ & 11.0\\
    \end{tabular}
    \newline
    \footnotesize{$(*)$optimization problem cannot be solved.}
  \end{center}
\end{table}

\begin{table}[h!]
  \begin{center}
    \caption{Response model evaluation on synthetic data 2 ($n_1=3, n_2=5, n_3=7$)}
    \label{tab:table2}
    \begin{tabular}{c|c|c|c} 
      &\textbf{DNN} & \textbf{Lattice} & \textbf{DIPN}\\
      \hline
      LogLoss & 0.8189 & 0.6267 & 0.5822 \\
      AUC-ROC & 0.7582 & 0.7579 & 0.7618 \\
      RPR(lower is better) & 0.40 & 0.00 & 0.00\\
      EPR(lower is better) & 0 & 0.05 & 0.00\\
      MLSS(lower is better) & 0.33 & 0.01 & 0.01\\
      Future response & 0.625 & 0.685 & 0.694\\
      Future cost($constraint=11.0$) & 7.5$(*)$ & 11.0 & 10.9\\
    \end{tabular}
    \newline
    \footnotesize{$(*)$optimization problem cannot be solved.}
  \end{center}
\end{table}


\begin{table}[h!]
  \begin{center}
    \caption{Response model evaluation on Production dataset}
    \label{tab:table3}
    \begin{tabular}{c|c|c|c} 
      &\textbf{DNN} & \textbf{Lattice} & \textbf{DIPN}\\
      \hline
      LogLoss & 0.2240 & 0.2441 & 0.2189 \\
      AUC-ROC & 0.7623 & 0.5000 & 0.7722 \\
      RPR(lower is better) & 0.28 & 0.00 & 0.00\\
      EPR(lower is better) & 0.00 & 1.00 & 0.08\\
      MLSS(lower is better) & 0.06 & 0.00 & 0.01\\
      Future response & 0.020 & 0.000 & 0.020\\
      Future cost($constraint=0.05$) & 0.050 & 0.000$(*)$ & 0.048\\
    \end{tabular}
     \newline
    \footnotesize{$(*)$optimization problem cannot be solved.}
  \end{center}
\end{table}

\subsection{Online Results}

We deployed our solution at Alipay and evaluated it in multiple marketing campaigns using A/B tests. Our solution consistently showed better performance than baselines, and was eventually deployed to all users. We show A/B test results for three marketing campaigns for HuaBei. HuaBei is a online micro-loan service launched by Ant Financial. It has about 300 million users according to public data. The incentive is electronic cash voucher, with usability subject to different terms in different campaigns. These campaigns include:
\begin{itemize}
\item Preferred Payment: the voucher can be cashed if a user set Huabei as the default payment method in the mobile application of Alipay.
\item New User: the voucher can be cashed if a user activates the Huabei service.
\item User Churn Intervention: the voucher can be cashed when a user uses Huabei to make a purchase.
\end{itemize}

All these campaigns have their corresponding business targets strongly correlated with voucher usage, so we use the usage rates as well as the monetary costs as evaluation metrics. \autoref{tab:table4} shows the relative improvements of DIPN compared to DNN model baselines. DIPN models use $30\%$ traffic in each experiment. 95 percentile confidence intervals are shown in square brackets. In all of the experiments, costs were significantly reduced. In two experiments, average voucher use rates were significantly increased.

\begin{table}[h!]
  \begin{center}
    \caption{Online Results}
    \label{tab:table4}
    \begin{tabular}{p{0.08\textwidth}|p{0.16\textwidth}|p{0.15\textwidth}} 
      &\textbf{Cost (\%)}  & \textbf{Usage Rate (\%)} \\
      \hline
      Payment Preferred & -6.05\%[-8.63\%, -3.47\%] & 1.86\%[-0.31\%, 4.04\%] \\
      New User Gift & -8.58\%[-10.51\%, -6.66\%] & 5.20\%[3.16\%, 7.23\%]\\
      User Churn Intervention & -9.42\%[-12.99\%, -5.84\%] & 8.45\%[4.53\%, 12.37\%]\\
    \end{tabular}
     \newline
  \end{center}
\end{table}

\section{Conclusion}
We focus on the problem of massive-scale personalized promotion in the internet industry. Such promotions typically require maximizing total user responses with limited budget. The decision variables for solving such problems are the incentive associated with the promotion.

We propose a two-step framework for solving such personalized promotion problems. In the first step, each user's response to each incentive level is predicted, and stored as coefficients for the second step. Predicting all these coefficients for many users is a counterfactual prediction problem. We recommend using randomized promotion policy or appropriate causal inference techniques such as IPS to process training data. To deal with data sparsity and noise, we designed a neural network architecture that incorporate our prior knowledge: (1) users' expected response monotonically increases with promotion incentive, and (2) the incentive-response curves are smooth. The proposed neural network, DIPN, ensures monotonicity and regularizes the magnitude of the first order derivative change. In the second step, an LP optimization problem can be formulated with the coefficients computed in the first step. We discussed its variants including (1) optimizing with more than one constraints, (2) supporting the constraint on average value, and (3) making decision for unseen users.

In experiments on synthetic datasets, we compared three algorithms: DNN, Deep Lattice Network, and our proposed DIPN. We show that DIPN has better performance in terms of data fitting, constraint violation, and promotion decisions. We also conducted a online experiment in one of Alipay's promotion campaign, in which user engagement was the desired response. DIPN got 6.05\% budget saving without losing user engagement, compared to DNN.

There are many possible extensions to the proposed framework. For example, the promotion response modeling can adapt to any causal inference techniques besides IPS\cite{dudik2011doubly}, for example causal embedding \cite{bonner2018causal} and instrumental variable\cite{hartford2016counterfactual}. Also the optimization formulation can be changed according to business requirements, as long as the computational complexity can be handled.



%

%
\bibliographystyle{ACM-Reference-Format}
\bibliography{pnp}

\end{document}